\documentclass[10pt]{article}
\usepackage{authblk}
\usepackage{blindtext}

\usepackage[utf8]{inputenc} 
\usepackage[T1]{fontenc}    
\usepackage{hyperref}       
\usepackage{url}            
\usepackage{booktabs}       
\usepackage{amsfonts}       
\usepackage{nicefrac}       

\usepackage{verbatim}

\usepackage{multirow}
\usepackage{amstext}
\usepackage{amsmath}
\usepackage{amsthm}
\usepackage{amssymb}
\usepackage{bm}
\usepackage{wrapfig}
\usepackage{amsfonts, amsmath}
\usepackage[utf8]{inputenc}
\usepackage{graphicx}
\usepackage{bbm, comment}
\usepackage{color}
\usepackage{float}
\usepackage{wrapfig}
\usepackage{url}
\usepackage{MnSymbol,wasysym,marvosym,ifsym}
\usepackage[T1]{fontenc}
\usepackage[inline]{enumitem}

\usepackage{autobreak}
\usepackage{microtype}      
\usepackage{booktabs} 
\usepackage[ruled]{algorithm2e} 
\usepackage[tight]{subfigure}
\usepackage{threeparttable}
\usepackage[utf8]{inputenc}
\usepackage{graphicx} 
\usepackage{float} 
\usepackage{pgfplots}
\usepackage{caption}
\usepackage{graphicx}

\usepackage{geometry}
\usepackage[numbers]{natbib}

\newcommand{\E}{\mathbb{E}}

\newtheorem{theorem}{Theorem}[section]
\newtheorem{lemma}[theorem]{Lemma}

\newtheorem{assumption}[theorem]{Assumption}

\newtheorem{example}[theorem]{Example}
\newtheorem{proposition}[theorem]{Proposition}
\newtheorem{definition}[theorem]{Definition}

\usepackage{xcolor}

\begin{document}

\title{Spot Check Equivalence: an Interpretable Metric for Information Elicitation Mechanisms}

\author[1]{Shengwei Xu\thanks{Corresponding Author}$^,$\thanks{Supported by United States National Science Foundation award number~2313137 and 2007256}$^,$}
\author[1]{Yichi Zhang$^{\dag,}$}
\author[1]{Paul Resnick}
\author[1]{Grant Schoenebeck$^{\dag,}$}

\affil[1]{School of Information, University of Michigan}
\affil[1]{\texttt{\{shengwei, yichiz, presnick, schoeneb\}@umich.edu}}

\maketitle

\begin{abstract}

Because high-quality data is like oxygen for AI systems, effectively eliciting information from crowdsourcing workers has become a first-order problem for developing high-performance machine learning algorithms.  Two prevalent paradigms, spot-checking and peer prediction, enable the design of mechanisms to evaluate and incentivize high-quality data from human labelers. So far, at least three metrics have been proposed to compare the performances of these techniques \citep{zhang2022high,gao2016incentivizing,burrell2021measurement}. However, different metrics lead to divergent and even contradictory results in various contexts. In this paper, we harmonize these divergent stories, showing that two of these metrics are actually the same within certain contexts and explain the divergence of the third. Moreover, we unify these different contexts by introducing  \textit{Spot Check Equivalence}, which offers an interpretable metric for the effectiveness of a peer prediction mechanism. Finally, we present two approaches to compute spot check equivalence in various contexts, where simulation results verify the effectiveness of our proposed metric.

\end{abstract}

\section{Introduction}\label{sec:intro}

Eliciting precise and valuable information from individuals is becoming paramount, especially with the rising demands for data labeling in the realms of AI and machine learning. Recent advancements, such as Large Language Models (LLMs), have proven the value of high-quality human-labeled data. For example, Meta is estimated to have invested upward of 25 million dollars in collecting preference data from human labelers to align Llama 2 with human preferences \citep{lambert2023}.  This raises a pressing question: How can human agents be incentivized to provide high-quality information. E.g., without the proper incentives,  human labelers for LLM alignment may not exert effort to distinguish between truthful LLM responses and merely authoritative-sounding responses (i.e.\ hallucinations), even when truthfulness is important for the task at hand. 

Research from Amazon reveals that monetary compensation is the principal motivator for Amazon Mechanical Turk workers \cite{chen2011opportunities}, and indeed the primary solution is to monetarily reward agents in exchange for effortful and truthful labeling. Two distinct compensation strategies, \textit{spot-checking}~\cite{jurca2005enforcing} and \textit{peer prediction}~\cite{miller2005eliciting}, each rate the quality of user feedback with a score.  This score can then be transformed into a payment for an agent.

Spot-checking mechanisms reward agents by comparing their reports with the ground truth on a small fraction of gold standard questions. When the ground truth information is expensive or even infeasible to obtain, peer prediction mechanisms are proposed, which reward an agent based on the correlation between her reports and the reports of other agents. 

To understand and compare the performance of mechanisms developed from these paradigms, there is a need for standard metrics similar to accuracy, recall, and F1 score used in supervised learning.  Notice that all these metrics range from 0 to 1 where 1 is good/perfect and 0 is bad.  While several studies have proposed methods for comparing these mechanisms \citep{zhang2022high,burrell2021measurement,gao2016incentivizing}, there remains a conspicuous gap for both a unified understanding of how these metrics relate, and, if possible, a unified interpretable metric. 

To this end, we introduce the concept of \textit{Spot Check Equivalence} (SCE), which uses a spot-checking mechanism as a benchmark to quantify the \emph{motivational proficiency} of an arbitrary incentive mechanism. As introduced in \citet{zhang2022high}, the motivational proficiency is the minimum cost of budget to induce a desired effort level in a symmetric equilibrium. Then, a SCE of 1 will indicate that a mechanism does as well as a certain spot-checking mechanism does when the spot-checking mechanism has access to the ground truth of every task.  A SCE of 0 will indicate that a mechanism does as well as this same spot-checking mechanism when it has no access to the ground truth of any task (essentially, it is paying agents randomly). Note that accessing the ground truth might be costly, e.g., hiring an expert to get the ground truth might be much more expensive than hiring several non-expert crowd workers. Thus, SCE can quantify the considerable cost savings that might be achieved by employing a peer prediction mechanism over a straightforward spot-checking mechanism.

By sufficiently harshly punishing the agents for the checked low-quality reports, spot-checking mechanisms can effectively motivate effort, even when only a small fraction of the tasks are checked. However, in most real applications, the payoff should be non-negative which precludes this approach. 

\citet{gao2016incentivizing, GAO2019618} study a peer grading setting where agents are modeled as having a binary choice for the effort to exert: low versus high.   In their setting, the goal is to minimize the fraction of random questions that must be spot-checked to incentivize agents to exert high effort (make choosing high effort a Nash equilibrium). They find that, in their model, combining spot-checking with peer prediction does not help reduce the spot-checking ratio required to achieve the desired incentive properties, i.e.~peer prediction makes things worse. However, their results rest on several assumptions, which we will discuss later in Section~\ref{sec:dis}. 

\citet{burrell2021measurement} propose a metric called \textit{Measurement Integrity} to quantify the ex-post fairness of a peer prediction mechanism. Mechanisms with high Measurement Integrity can produce payments that are strongly correlated with the quality of the agents' reports. Their motivation and definitions look beyond incentives to fairness. They do not study spot-checking mechanisms, but it is clear that the more agents are spot-checked the more accurately their scores will reflect their true quality.  For example, with ground truth for all the tasks, an agent's score could exactly reflect the true quality of her responses.  Moreover, they model continuous effort but do not establish a clear link between Measurement Integrity and the ability to incentivize effort.

\citet{zhang2022high} study incentivizing effort in a crowdsourcing setting where agents can choose their effort from a continuum. Their goal is to maximize the motivational proficiency by rescaling the scores output by the incentive mechanism into practical payments. They suggest a tournament-based payment scheme: first rank the agents based on their scores output by an incentive mechanism, and then pay a predetermined reward for each ranking. Under the tournament setting, they further propose a sufficient statistic of a mechanism's motivational proficiency, called the \textit{Sensitivity}. Intuitively, the Sensitivity measures how responsive a score is to changes in an agent's effort. For example, a spot-checking mechanism that checks a larger fraction of tasks is more sensitive to changes in an agent's effort (intuitively, it has more chances to detect a change), and thus has a larger motivational proficiency. However, there is a lack of discussions on how to estimate Sensitivity in practice.

An apparent contradiction arises. \citet{zhang2022high} empirically show that when agents exert a reasonably high effort, peer prediction mechanisms have Sensitivity competitive with spot-checking mechanisms that randomly check 20\% of the tasks. However, the aforementioned implication of \citet{gao2016incentivizing, GAO2019618} would seem to predict that the spot-checking mechanisms are always superior to peer-prediction mechanisms.

\paragraph{Our contributions.} 

First, we propose Spot Check Equivalence which uses the equivalent spot-checking ratio as an interpretable way to measure an information elicitation mechanism's performance under specified information elicitation contexts. We study the Spot Check Equivalence based on Measurement Integrity and Sensitivity, and demonstrate its effectiveness as a metric for motivational proficiency both theoretically and empirically. 

Second, we unify Measurement Integrity (the metric for ex-post fairness) and Sensitivity (the metric that serves as a proxy for motivational proficiency). In particular, we prove that Spot Check Equivalence based on Measurement Integrity and Sensitivity are sometimes exactly the same. We also show why these results differ so much from \citet{gao2016incentivizing, GAO2019618}, and thus refute, or at least qualify, the titular statement that ``Peer-prediction makes things worse."  

Third, we present two approaches to compute Spot Check Equivalence, which are suitable for settings with and without ground truth data, respectively. Our method enables the comparison of the motivational proficiency of different mechanisms across various information elicitation contexts. Furthermore, our simulation results show that both approaches result in similar estimations of SCE, which implies the robustness of our methods.

\section{Model}\label{sec:model}

\begin{figure}[!ht]
    \centering
    \includegraphics[width=1\linewidth]{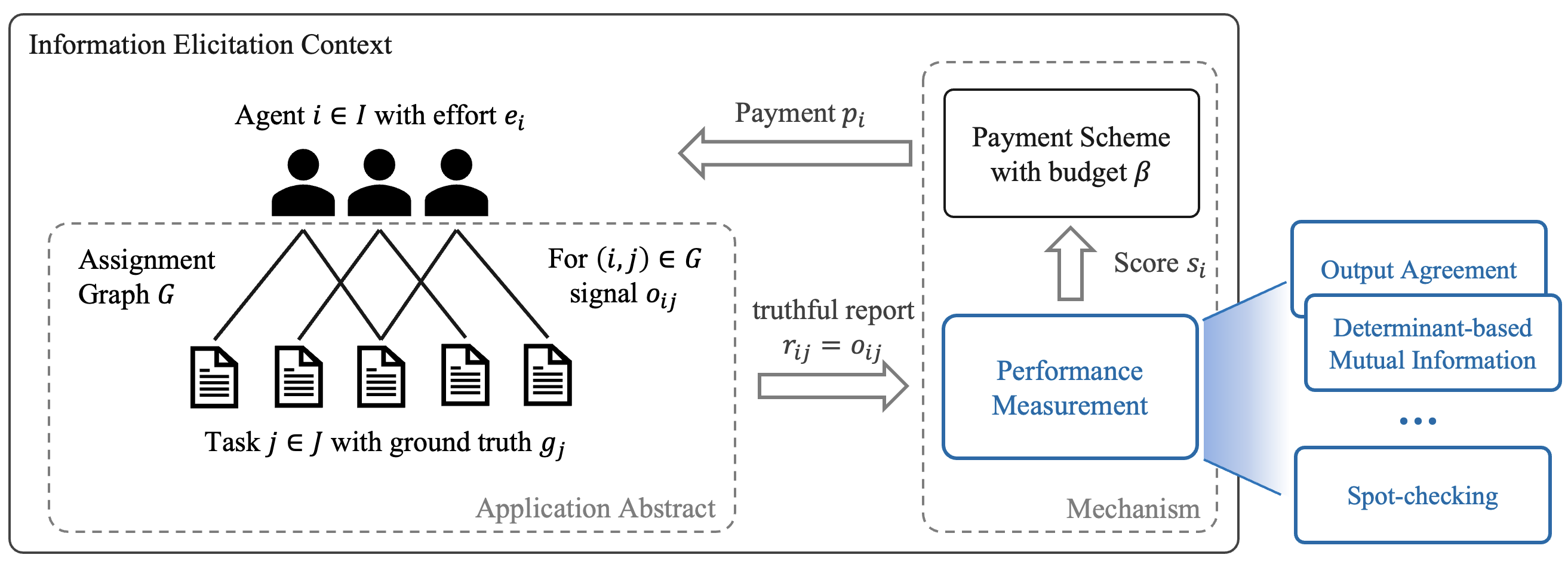}
    \caption{Information Elicitation Context}
    \label{fig:info-elicit-context}
\end{figure}

In this section, we will give a formal definition of the information elicitation context (Figure~\ref{fig:info-elicit-context}), and then formally define the Spot Check Equivalence.

\subsection{Information Elicitation Context}

Formally, as shown in Figure~\ref{fig:info-elicit-context}, an information elicitation context (IEC) is defined as a tuple: 
\[\text{Information Elicitation Context ($IEC$)}:=(Agent,App,Mech)\]
where $Agent=(I,c,\mathbf{e})$ represents the agents and their properties, $App=(J,\mathcal{GT},\omega,\Sigma,D)$ represents an information elicitation application abstraction, and $Mech=(M,P)$ represents a mechanism. We assume that the information elicitation context is common knowledge for all the agents.

\paragraph{Agent}
In $Agent=(I,c,\mathbf{e})$: $I$ is the set of agents; $\mathbf{e}=[e_i]_{i\in I} \in [0,1]^{|I|}$ represents all agents' effort levels.\; the cost function  $c: [0,1] \rightarrow \mathbb{R^+} \cup \{0\}$ maps an effort level to a non-negative, increasing, and convex cost.  Notice agents are homogeneous and share the same cost function.

\paragraph{Application Abstraction} 

$App=(J,\mathcal{GT},\omega,\Sigma,D)$ comprises the task set $J$, the ground truth space $\mathcal{GT}$, the prior of the ground truth $\omega = \Delta_{\mathcal{GT}}$, the signal space $\Sigma$, and the data-generating process $D = (D_{assign},D_{signal})$.

$D_{assign}$ describes how the tasks are assigned to the agents.
\[D_{assign}: \Delta_{\mathcal{G}}\]
where $\mathcal{G}$ represents the space over $G$, and $G=(I\cup J,E_G)$ represents a bipartite graph between $I$ and $J$, indicating how the tasks are assigned to the agents.

$D_{signal}$ describes how the signals are generated: the distribution of agent $i$'s signal on task $j$ conditioned on the effort $e_i\in [0,1]$ and task $j$'s ground truth $g_j\in\mathcal{GT}$, given the edge $(i,j)\in E_G$:
\[D_{signal}: [0,1]\times  \mathcal{GT} \rightarrow \Delta\Sigma\]

\paragraph{Agents' Report} We assume agents truthfully report the signals they obtain from the application abstraction conditioned on their effort levels. Further discussion will be provided in Appendix~\ref{subsec:truthful-asp}. And we denote the agent $i$'s report on task $j$ as $r_{ij} \in \Sigma$.

\paragraph{Application Instance} With the specified $Agent=(I,c,\mathbf{e})$ and $App=(J,\mathcal{GT},\omega,\Sigma,D)$, we can generate an instance representing a realized information elicitation application:
\begin{itemize}
    \item For the given $I$, $J$, we sample an assignment graph $G$ according to $D_{assign}$.
    \item For each task $j\in J$, we independently sample its ground truth $g_j$ from the prior $\omega$.
    \item For each pair $(i,j)\in E_G$, we independently sample agent $i$'s signal on task $j$ from the distribution $D_{signal}(e_i,g_j)$, denoted as $o_{ij}\in \Sigma$.
    \item For each pair $(i,j)\in E_G$, as we assumed, the agent $i$'s report $r_{ij} = o_{ij}$.
\end{itemize}

The mechanism, introduced presently, takes the application instance as input.

\paragraph{Performance Measurement}

The performance measurement $M$ is a component of the mechanism $Mech=(M,P)$. It maps the agents' reports to their performance scores. Formally,
\[\text{(Peer Prediction)}~M: \Sigma^{|E_G|} \rightarrow_{random} \mathbb{R}^{|I|}\]
\[\text{(Spot-checking)}~M:  \Sigma^{|E_G|} \times \mathcal{GT}^{|J_\text{checked}|} \rightarrow_{random} \mathbb{R}^{|I|}\]
Note that the spot-checking performance measurement can access the ground truth of the checked tasks $J_\text{checked} \subseteq J$.

We use $s_i \in \mathbb{R}$ to denote agent $i$'s score, and $\mathbf{s}=[s_i]_{i\in I}$ to denote the vector of all agents' scores.

\paragraph{Payment Scheme}
The payment scheme $P$ is the other component of the mechanism. It maps the agents' performance scores to the payoffs, which are directly related to their utilities. Formally,
\[\text{(Payment Scheme)}~P: \mathbb{R}^{|I|} \rightarrow {\left(\mathbb{R}^* \cup \{0\}\right)}^{|I|}\]

We use $p_i \in \mathbb{R}^* \cup \{0\}$ to denote payoff of agent $i$, and $\beta = \sum_{i\in I} p_i$ to denote the total payment among all the agents. And we denote the vector of all the agents' payoffs as $\mathbf{p}=[p_i]_{i\in I}$.

For intuition, we give two examples of payment schemes:

\begin{definition}[Linear Payment Scheme]\label{def:linear-pay}
    A linear payment scheme pays a payoff $p_i=a\cdot s_i+b$ to each agent $i$, where $a,b$ are the constant parameters.
\end{definition}

\begin{definition}[Tournament Payment Scheme]\label{def:tournament-pay}
    A tournament payment scheme first ranks the agents according to their scores and then pays the $i$-th ranked agent $\hat{p_i}$, where $\hat{p_1},\hat{p_2},...,\hat{p_{|I|}}$ are constant parameters that monotonically decrease, i.e., $\hat{p_i}\geq \hat{p_{i'}}$ when $i\leq i'$. Without loss of generality, we assume $s_1\geq s_2\geq ... \geq s_{|I|}$, and thus $p_i=\hat{p_i}$. 
\end{definition}

\paragraph{Report quality.} Given an instance, we can define the quality of an agent's report. The quality function $Q$ for one report is a deterministic loss function:
\[\text{(Quality Function)}~Q: \Sigma \times \mathcal{GT} \rightarrow \mathbb{R}\]
Agent $i$'s overall report quality $q_i$ is defined as the average of her reports' qualities, i.e. $q_i=\sum_{j|(i,j)\in G} Q(r_{ij},g_j)$. We denote the vector of all the agents' qualities as $\mathbf{q}=[q_i]_{i\in I}$. 

\begin{figure}[!ht]
    \centering
    \includegraphics[width=0.7\linewidth]{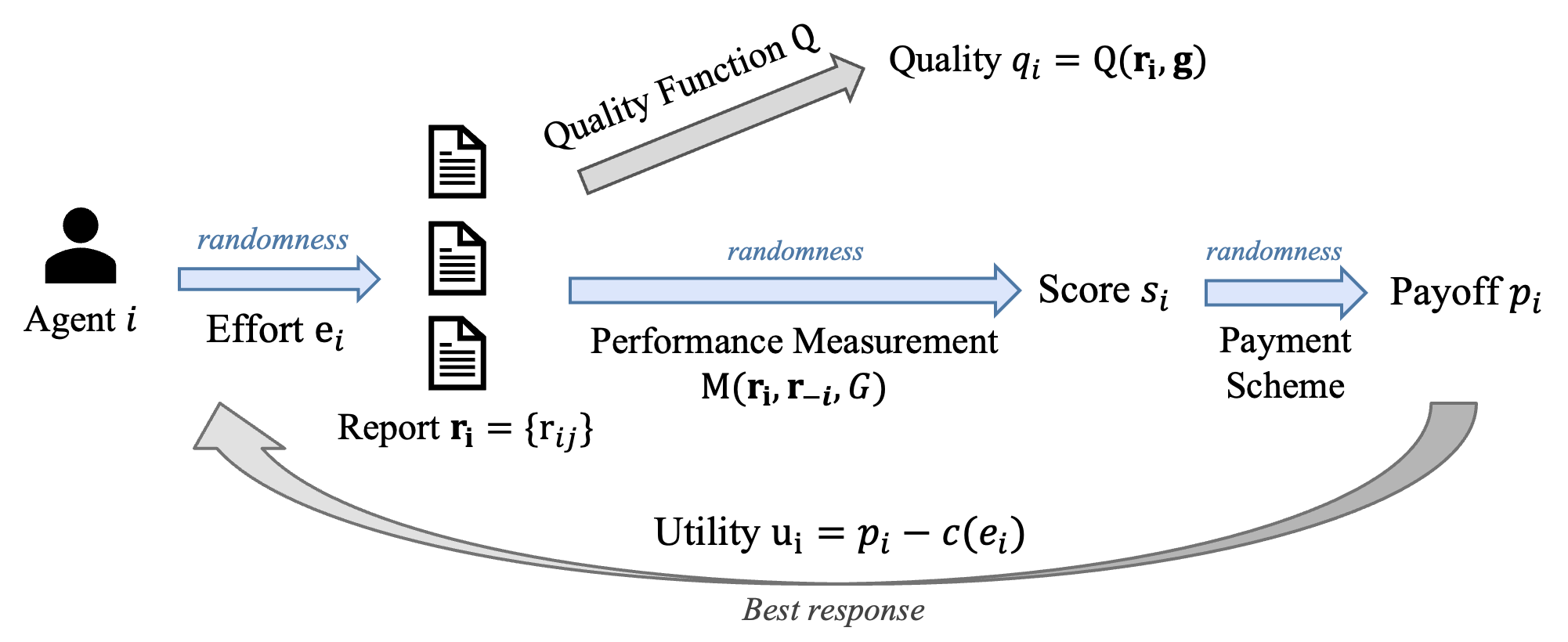}
    \caption{An Agent's Perspective of an Information Elicitation Context}
    \label{fig:effort-report-score}
\end{figure}

\paragraph{Equilibrium}

We assume that the agents choose their effort level according to the following equilibrium.

\begin{definition}[Symmetric local equilibrium]\label{def:symmetric-eq}
Given an IEC where all the agents exert effort $e_i=\xi$ and $\E[\sum_{i\in I} p_i]\geq |I|\cdot c(\xi)$ (Individual Rationality is satisfied), we say it is a symmetric local equilibrium if the derivative of every agent's utility is $0$, i.e. \[\frac{\partial}{\partial e_i}u(e_i,e_{-i}=\xi)|_{e_i=\xi}=0\]
\end{definition}

Note that, at this equilibrium, $\frac{\partial}{\partial e_i}u(e_i,e_{-i}=\xi)|_{e_i=\xi}=0$ is a necessary condition for $\xi$ being a Nash equilibrium. \citet{zhang2022high} show empirical evidence that a local equilibrium is very likely to be a Nash equilibrium in the model we mainly discuss in our paper. In the rest of our paper, our discussion will focus on this equilibrium.

\subsection{Motivational Proficiency}

The motivational proficiency of a component (a mechanism, a performance measurement, or a payment scheme) within an information elicitation context $IEC$ represents its ability to incentivize effort. To quantify it, we fix all the other components of the $IEC$ and quantify the expected total payment for eliciting a fixed effort level at the equilibrium (Definition~\ref{def:symmetric-eq}), lower expected total payment implies higher motivational proficiency.

\begin{definition}[Motivational proficiency] \label{def:incentive-effort}
We define the motivational proficiency of a component ($Mech$, $M$, or $P$) within an information elicitation context $IEC$ where all the agents exert effort level $\xi$ as the negative expected total payment needed to realize the symmetric local equilibrium (Definition~\ref{def:symmetric-eq}) at effort level $\xi$ when substituting the component into the information elicitation context. 
\end{definition}

As we discussed in the introduction, \citet{zhang2022high} show that tournament payment schemes have higher motivational proficiency compared to linear payment schemes in certain settings where limited liability is needed. Therefore, in the following discussion, we will focus on the motivational proficiency of performance measurements in information elicitation contexts with a tournament payment scheme. In Section~\ref{sec:exp}, we quantify the total payment in an information elicitation context with tournament payment. 

We provide a simple example that illustrates the calculation of expected total payment in an information elicitation context with a winner-take-all tournament payment scheme. In the next section, we then demonstrate the Sensitivity and the Measurement Integrity in this example.

\begin{example} \label{eg:more-checking}
Consider a simple case where there are two agents (1 and 2) working on a large number of tasks. Each agent $i\in\{1,2\}$ can choose to exert a non-negative effort level $e_i$, which incurs a cost $c(e_i)$. And we use a quadratic cost function, $c(e_i) = e_i^2$, which is one of the simplest functions that satisfies all the properties of a cost function.

Since the signals are independent conditioning on an agent's effort level and the spot-checking performance measurement checks each task u.a.r., we use normal distributions to approximate the quality and performance score: Agent $i$'s report quality is $q_i \sim N(e_i,1)$, and the performance score is $s_i \sim N(q_i,\sigma^2)$, where $\sigma$ is monotonically decreasing with the spot-checking ratio. Consequently, we have that $s_i \sim N(e_i,\sigma^2 + 1)$

The expected utility of agent $i$ is \[u_i(e_i)=\Pr[i \text{ wins the tournament}]\cdot \operatorname{payoff} - c(e_i)\]

$\Pr[i \text{ wins the tournament}] = \Pr[s_i - s_{-i} \geq 0]$, where $-i$ denotes the other agent. 
Notice that $(s_i - s_{-i}) \sim N(e_i - e_{-i},2\sigma^2 + 2)$ because the difference of two normal random variables is also a normal random variable where the variances add. Therefore, we have \[\Pr[i \text{ wins the tournament}] = CDF_{N(e_{-i},2\sigma^2+2)}(e_i)\] where $CDF_{N(e_{-i},2\sigma^2+2)}$ is the cumulative distribution function of the normal distribution $N(e_{-i},2\sigma^2+2)$.

To achieve the symmetric local equilibrium at effort level $\xi$, we let the derivative of agent $i$'s expected utility at $e_i=\xi$ equal to 0. 
\begin{align*}
    \frac{\partial}{\partial \xi}u_i(\xi) = {pdf_{N(\xi,2\sigma^2+2)}(\xi)} \cdot \operatorname{payoff} - 2\xi = 0
\end{align*}
where ${pdf_{N(\xi,2\sigma^2+2)}}$ is the probability density function of normal distribution $N(\xi,2\sigma^2+2)$, which is the derivative of $\Pr[i \text{ wins the tournament}]$ at $e_i = \xi$. By simplification, we get, $pdf_{N(\xi,2\sigma^2+2)}(\xi) = \frac{1}{2\sqrt{\pi\cdot (\sigma^2+1)}}$. Solving for the payoff, we have
\begin{align*}
     \operatorname{payoff} = 4\xi \cdot \sqrt{\pi\cdot (\sigma^2+1)}
\end{align*}

In addition, we need to keep Individual Rationality, i.e. the expected utility for each agent should not be negative so that rational agents will not leave, which implies the total payoff to all agents should cover the total cost of all agents, i.e., $\operatorname{payoff} \geq 2c(\xi) = 2\xi^2$. Putting these together, the payoff should be
\[\operatorname{payoff}=\max\left(2\xi^2,4\sqrt{\pi\cdot (\sigma^2+1)}\cdot \xi\right)\]
\end{example}

In the above example, we note that, $\sigma$ monotonically decreases with the spot-checking ratio, and the total payoff monotonically increases with $\sigma$ when IR is not binding. Therefore, we can see that a higher spot-checking ratio leads to a lower total payment when IR is not binding, and consequently, a higher motivational proficiency.

\subsection{Measure of Performance Measurements}

In addition to motivational proficiency, there are other measures of performance measurements, as we discussed in the introduction, including Sensitivity \citep{zhang2022high} and Measurement Integrity \citep{burrell2021measurement}. We propose the general definition of a measure of performance measurement $M$.

\begin{definition}[Measure of Performance Measurement]\label{def:measure}
    A measure $f$ of performance measurement $M$ within information elicitation context $IEC$ maps the $IEC$ with $M$ to a real number, denoted as $f(IEC \leftarrow M)\in \mathbb{R}$, where the leftarrow means we apply $M$ in the information elicitation context $IEC$.
\end{definition}

We now show the two examples of measure $f$, Sensitivity \citep{zhang2022high} and Measurement Integrity \citep{burrell2021measurement}.

\paragraph{Sensitivity} \citet{zhang2022high} propose the Sensitivity as a proxy of the motivational proficiency of a performance measurement. They show that the motivational proficiency highly depends on the Sensitivity (Definition~\ref{def:Sensitivity}, Lemma~\ref{lem:Sensitivity}), which measures how an agent's performance score changes when she deviates from the equilibrium effort level $\xi$. When all other agents exert effort $\xi$, we denote the mean of agent $i$'s score as $\mu_s(e_i)$, and the standard deviation as $\sigma_s(e_i)$.

\begin{definition}[Sensitivity \cite{zhang2022high}]\label{def:Sensitivity} The Sensitivity of a performance measurement within an information elicitation context $IEC$ at equilibrium effort level $\xi$ is defined as 
\[\operatorname{Sensitivity}(IEC\leftarrow M) = \delta(\xi) =\frac{\frac{\partial }{\partial e_i}\mu_{s}(e_i)|_{e_i=\xi}}{\sigma_s(\xi)}\]
\end{definition}

\begin{lemma}[Proposition 4.8 in \citet{zhang2022high}]\label{lem:Sensitivity}
If the agent $i$'s score $s_i$ follows a normal distribution $N(\mu_s(e_i),\sigma_s(e_i)^2)$, the expected total payment to elicit effort $\xi$ will (weakly) decrease in the Sensitivity $\delta(\xi)$ in a specific information elicitation context with any tournament payment scheme.
\end{lemma}

To illustrate, we show the Sensitivity $\delta(\xi)$ in Example~\ref{eg:more-checking}. 
\[\delta(\xi)=\frac{\frac{\partial }{\partial e_i}\mu_{s}(e_i)|_{e_i=\xi}}{\sigma_s(\xi)} = \frac{1}{\sqrt{\sigma^2+1}}\]

We find that the total payment in Example~\ref{eg:more-checking}, $\max\left(2\xi^2,4\sqrt{\pi\cdot (\sigma^2+1)}\cdot \xi\right)$, is inversely proportional to the Sensitivity when IR is not binding, which aligns with the Lemma~\ref{lem:Sensitivity}.

\paragraph{Measurement Integrity} To measure the ex-post fairness of a performance measurement, \citet{burrell2021measurement} propose the Measurement Integrity, which is defined as the expected correlation between the quality of the agents' reports and their performance scores. 

\begin{definition}[Measurement Integrity]\label{def:mi}
Formally, the Measurement Integrity of a performance measurement $M$ with respect to  a quality function $Q$ and a correlation function $\operatorname{corr}$, within an information elicitation context $IEC$ is
\[
\underset{Q, \operatorname{corr}}{\operatorname{MI}}(IEC\leftarrow M)=\mathbb{E}_{IEC}\left[\operatorname{corr}\left(\mathbf{s}, \mathbf{q}\right)\right]
\]

\end{definition}

In Example~\ref{eg:more-checking}, we can see that the motivational proficiency highly depends on how an agent's performance score correlates with her report quality. In the example, the Pearson correlation coefficient between the agent i's report quality and score is as follows.
\begin{align*}
    \rho(q_i,s_i) = \frac{\E[q_i\cdot s_i]-\E[q_i]\E[s_i]}{\sigma_{q_i}\cdot \sigma_{s_i}} = \frac{1}{\sqrt{\sigma^2+1}}
\end{align*}

Note that the total payment is also inversely proportional to the Pearson correlation coefficient! And the Sensitivity has the same form as the correlation in this example.

This example intuitively shows that the correlation between the agents' report qualities and scores can be a proxy for a performance measurement's motivational proficiency. And both the report quality and score are accessible in real data.

Therefore, we can employ both Sensitivity \citep{zhang2022high} and Measurement Integrity \citep{burrell2021measurement} as our proxy for motivational proficiency. In Section~\ref{sec:theory}, we further formally unify these two proxies within our model.

\subsection{Using Spot-checking as a Reference: Spot Check Equivalence}

Even though we have metrics which can compare two performance measurements, there is still a need for a metric for a performance measurement whose value is per se. Our idea is to take any metric of performance measurements, $f$, and convert it to an interpretable metric as follows: instead of using the actual value of $f$ applied to some performance measurement, instead use the checking ratio of the spot-checking performance measurement who, when also evaluated by $f$, yields a value equivalent to that of the performance measurement. 

First, we adopt the definition of spot-checking performance measurement from \citet{gao2016incentivizing}, and assume it can access the ground truth, thus, it is an idealized spot-checking. Given that we exclusively focus on this particular spot-checking, we might omit the term 'idealized' in subsequent discussions for brevity.

\begin{definition}[Spot-checking performance measurement (idealized)] \label{def:sc}
An (idealized) spot-checking performance measurement is denoted as a tuple $SC:=(X, S, C)$, which checks $X$ percent of all the tasks u.a.r.; then scores the agent $i$ with $S(r_{ij},g_j)$ for each checked task $j$, and score $C \in \mathbb{R}$ for each of the unchecked tasks.
\end{definition}

Intuitively, a higher checking ratio leads to less noise, so high effort is easier to notice, which is beneficial for both motivational proficiency and ex-post fairness. Thus, we can use the equivalent spot-checking ratio as a metric for both motivational proficiency and ex-post fairness of a performance measurement. 

Formally, we can define the Spot Check Equivalence as follows. 

\begin{definition}[Spot Check Equivalence]\label{def:gr}
    
For a performance measurement $M$ within an information elicitation context $IEC=(Agent,App,Mech)$, at the symmetric local equilibrium $\mathbf{e}=\xi$, given a measure of performance measurements $f$ (Definition~\ref{def:measure}), we define the Spot Check Equivalence $SCE$ of $M$, with respect to a spot-checking mechanism $SC$ as
\begin{align*}
    SCE(IEC \leftarrow M) = \operatorname{sup}_{X} \{ f(IEC\leftarrow M) \geq f(IEC\leftarrow SC(X,S,C))\}
\end{align*}

\end{definition}
In our following discussion, to make the spot-checking mechanism $SC(X,S,C)$ non-trivial\footnote{An example of a trivial spot-checking mechanism is $S(r_{ij},g_j) \equiv 0$ and $C \equiv 0$.}, we use a quality function to score the agents for checked tasks, i.e. $S=Q$. And since we will apply a payment scheme after the performance measurement, the value of the constant score $C$ for unchecked tasks does not matter, thus, we set $C=0$. We then use $SC(X)$ to denote $SC(X,S=Q,C=0)$.

In the next section, we will theoretically prove the unification of Sensitivity and Measurement Integrity in certain settings, and consequently, we can show that the Spot Check Equivalence based on the Sensitivity or Measurement Integrity can be used as an interpretable metric for the motivation proficiency of an information elicitation performance measurement. We will show more empirical evidence in our agent-based model simulations (Section~\ref{sec:exp}) when relaxing the theoretical assumptions.

\section{Unification of Sensitivity and Measurement Integrity}\label{sec:theory}

In this section, we formally prove that the Spot Check Equivalence based on Sensitivity or Measurement Integrity can be used as a proxy for Spot Check Equivalence based on motivational proficiency in certain settings. Given that computing these three measures has different requirements, the unification allows us to compute Spot Check Equivalence and consequently, measure the motivational proficiency in more scenarios. 

We formally propose and prove the unification of Measurement Integrity and Sensitivity in our main theorem (Theorem~\ref{thm:main}). Recall that \citet{zhang2022high} have shown that, within certain settings, high Sensitivity leads to low expected total payment (Lemma~\ref{lem:Sensitivity}) when applying a tournament payment scheme. Putting these together, we  have that the Spot Check Equivalence based on motivational proficiency, Sensitivity, and Measurement Integrity are equal.

\begin{figure}[!ht]
    \centering
    \includegraphics[width=0.6\linewidth]{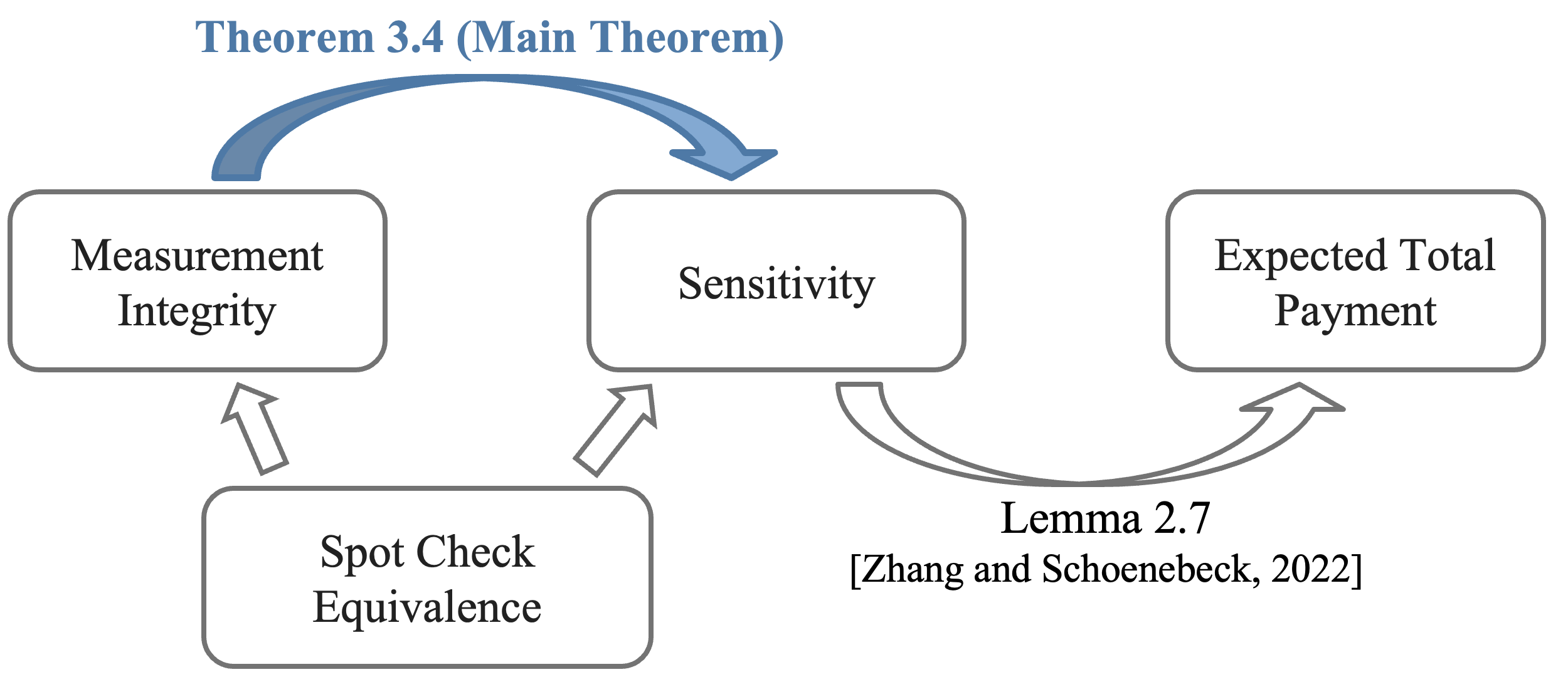}
    \caption{Theoretical Analysis Overview: Spot Check Equivalence based on Measurement Integrity, Sensitivity and motivational proficiency (negative expected total payment) are equivalent.}
    \label{fig:theorem-overview}
\end{figure}

Since the Sensitivity relies on the distribution of the performance score, we make the following assumptions about the distributions of the quantities in our model, similar to prior work~\citep{zhang2022high}. 

\begin{assumption}[The Gaussian assumption for the quality]\label{asp:gaussian-quality}
Given effort level $e_i$, the quality $q_i$ follows a normal distribution $N(e_i,\sigma_q(e_i)^2)$. And we further assume that $\sigma_q'(e_i)<<\sigma_q(e_i)$.
\end{assumption}

\begin{assumption}[The Gaussian assumption for the score]\label{asp:gaussian-score}
Given the report quality $q_i$, the score $s_i$ follows a normal distribution $N(\mu_{s|q}(q_i),\sigma_{s|q}(q_i)^2)$.
\end{assumption}

\citet{zhang2022high} show empirical evidence that the performance score roughly follows a normal distribution, here, we further make this assumption for the quality, a quantity that did not exist in the prior work~\citep{zhang2022high}. When the number of agents and the number of tasks per agent grows large, it is reasonable to assume the quality is normal distributed because it can typically be seen as the sum of many i.i.d. random variables.

Note $\mu_{s|q}$ is different from the $\mu_s$ in the definition of Sensitivity, in particular,
\begin{align*}
    \mu_s(e) = \int_q \mu_{s|q}(q)  \Pr[q|e] \operatorname{dq} 
\end{align*}

\begin{assumption}[Independence assumption for scores]\label{asp:score-independence}
When all agents have the same effort level $e_i=\xi$ and the number of agents $|I|$ goes to infinity, the agents' performance scores are independent. 
\end{assumption}

Note that different agents' qualities $q_i$ are independent. When we apply a spot-checking performance measurement, the scores are naturally independent. When we apply a peer-sensitive performance measurement (e.g. peer prediction), an agent's performance score depends on her report quality as well as other agents' report quality. However, as the number of agents grows large, the amount that one agent's reports can impact another agents' scores will go to zero.   Thus, we assume that the scores are independent when the number of agents goes to infinity (Assumption~\ref{asp:score-independence}).

\begin{theorem}[Main Theorem]\label{thm:main}
For a given performance measurement $M$ within an information elicitation context $IEC$ where every agent exerts effort level $\xi$, when Assumption~\ref{asp:gaussian-quality} \ref{asp:gaussian-score} and \ref{asp:score-independence} are satisfied, there exists a linear bijection between the ${\operatorname{ MI }}_{Q, \operatorname{corr}}(IEC\leftarrow M)$ and the Sensitivity $\delta(\xi)$, where $\operatorname{corr}$ is the sample Pearson correlation coefficient and the number of agents goes to infinity.
\end{theorem}

The proof of Theorem~\ref{thm:main} is provided in Appendix~\ref{apx:theory}.

Note that our main theorem (Theorem~\ref{thm:main}) relies on Assumption~\ref{asp:gaussian-quality}, \ref{asp:gaussian-score}, and \ref{asp:score-independence}, thus, it is important to examine whether the unification of motivational proficiency, Sensitivity, and Measurement Integrity is still true with real scores calculated by various performance measurements. In the section~\ref{sec:exp}, we will demonstrate some positive evidence from our agent-based model experiment, where Assumption~\ref{asp:gaussian-quality}, \ref{asp:gaussian-score}, and \ref{asp:score-independence} are relaxed.

\section{Computing Spot Check Equivalence}

Given the unification of Sensitivity and Measurement Integrity, if we can compute the Sensitivity or Measurement Integrity of a performance measurement, we can get the Sensitivity-based or Measurement Integrity-based Spot Check Equivalence respectively, and consequently, we can get an interpretable metric for motivational proficiency.

If the measure $f(IEC\leftarrow SC(X))$ is monotonic with respect to $X$, we can use a binary search algorithm (Algorithm~\ref{alg:bisearch}) to compute the Spot Check Equivalence.

\SetKwProg{Fn}{Function}{:}{}
\SetKwFunction{FMain}{BinarySearchSCE}

\begin{algorithm}
\DontPrintSemicolon
\KwIn{Information Elicitation Context $IEC$, Performance Measurement $M$, step size $\epsilon$}
\KwOut{Spot Check Equivalence $SCE$}

\Fn{\FMain{$M, f$}}{
    $\text{low} = 0, \text{high} = \lfloor100/\epsilon\rfloor$\;
    \While{$\text{low} \leq \text{high}$}{
        $\text{mid} = \left\lfloor \frac{\text{low} + \text{high}}{2} \right\rfloor$\;
        \If{$f(IEC\leftarrow SC(X=\text{mid}*\epsilon)) < f(IEC\leftarrow M)$}{
            $\text{ans} = \text{mid}$\;
            $\text{low} = \text{mid} + 1$\;
        }
        \Else{
            $\text{high} = \text{mid} - 1$\;
        }
    }
    \KwRet{\text{ans}}\;
}
\caption{Binary Search algorithm for SCE}
\label{alg:bisearch}
\end{algorithm}

Furthermore, a linear combination of the two spot-checking ratios which has adjacent measure $f$ may be applied for a better approximation.
\[SCE = \text{ans} + \epsilon \cdot \frac{f(IEC\leftarrow M)-f(IEC\leftarrow SC(\text{ans}))}{f(IEC\leftarrow SC(\text{ans}+\epsilon))-f(IEC\leftarrow SC(\text{ans}))}\]

\subsection{Computation with Ground Truth}

We first propose a workflow to compute the Spot Check Equivalence with the ground truth of the tasks. The Spot Check Equivalence is like accuracy, recall, and F1 score in machine learning, which can only be calculated on training or testing datasets rather than real applications. Similarly, it is reasonable to create datasets to evaluate the information elicitation performance measurements, get a good sense of their motivational proficiency, and then decide which performance measurement to apply in real applications.

With the ground truth of the tasks, we can calculate the quality of the reports, and then, use the correlation between the agents' scores and qualities to estimate the measurement integrity of the performance measurement $M$ and $SC(X)$. 

Intuitively, as more tasks are checked, the score $s_i$ will be more correlated to the quality $q_i$. Our agent-based model experiment also shows that the Measurement Integrity monotonically increases with the spot-checking ratio (Section~\ref{subsec:MI-SEC}). Therefore, we can apply Algorithm~\ref{alg:bisearch} to estimate the Spot Check Equivalence based on Measurement Integrity.

\subsection{Computation without Ground Truth}

Considering the current lack of information elicitation dataset, we propose another method to estimate the Spot Check Equivalence without the ground truth.

Recall the definition of Sensitivity (Definition~\ref{def:Sensitivity}), both the derivative of the performance score $\frac{\partial }{\partial e_i}\mu_{s}(e_i)|_{e_i=\xi}$ and the standard deviation $\sigma_s(\xi)$ do not require access to the ground truth. The standard deviation $\sigma_s(\xi)$ can be estimated by the standard deviation of $\{s_i\}$ given that all the agents are homogeneous. However, in real data, to estimate $\frac{\partial }{\partial e_i}\mu_{s}(e_i)|_{e_i=\xi}$ is tricky because the score after deviating from $\xi$ is not accessible.

We adopt the idea of bootstrap sampling: we randomly select an agent $i$, and if we know how the effort impacts the report, we can randomly manipulate her report as an intentional degradation of her effort by some $\varepsilon$. We then compute the mean difference between all selected agents' scores before and after the manipulation, denoted as $\Delta \mu$. Note that the mean difference $\Delta \mu$ can be regarded as an estimation of $\varepsilon \cdot \frac{\partial }{\partial e_i}\mu_{s}(e_i)|_{e_i=\xi}$, and consequently, $\Delta \mu/ \sigma_s$ is proportional to the Sensitivity. For example, if decreasing the effort brings uniform noise, we get the Algorithm~\ref{alg:sce_no_gt}. In Section~\ref{subsec:exp-sensitivity}, we present evidence demonstrating that Algorithm~\ref{alg:sce_no_gt} works in our agent-based model simulation.

\begin{algorithm}
\DontPrintSemicolon
\KwIn{Information Elicitation Context $IEC$, Performance Measurement $M$, Iteration Times $T$}
\KwOut{Spot Check Equivalence $SCE$}
\Fn{Estimate~$\Delta\mu(IEC\leftarrow M)$}{
    $\Delta\mu = 0$\;
    \For{$t = 1$ \KwTo $T$}{
        Choose $i \in I$ u.a.r.\;
        Compute score $s_i$ with $M$\;
        \For{$(i,j) \in \mathcal{G}$}{
            \If{$\text{random}(0, 1)>\varepsilon$}{
                Choose $r_{ij} \in \mathcal{GT}$ u.a.r.\;
            }
        }
        Compute score $s_i'$ with $M$\;
        $\Delta\mu = \Delta\mu + (s_i' - s_i)/T$\;
    }
    \KwRet $\Delta\mu$\;
}
\BlankLine
$SCE =$ \FMain{$M, f=\Delta\mu/\sigma_s$}\;
\caption{Estimate SCE without Ground Truth}
\label{alg:sce_no_gt}
\end{algorithm}

\section{Effectiveness of the Spot Check Equivalence: Agent-Based Model Experiment}\label{sec:exp}
In this section, we present the results of agent-based model (ABM) experiments to evaluate the effectiveness of the \textit{Spot Check Equivalence} based on Measurement Integrity and Sensitivity in measuring the motivational proficiency, without the assumptions we made in our theoretical proof. We then further compare different peer prediction performance measurements with spot-checking performance measurements in different information elicitation contexts.

\subsection{Setup}

In this subsection, we introduce our agent-based model setup. According to the formal definition of the information elicitation context in Section~\ref{sec:model}, our agent-based model contains the following components:

\subsubsection{Agents}

We consider a population of $|I|=50$ agents.

\paragraph{Effort and Cost.} Each agent $i$ has an effort level $e_i \in [0, 1]$ with an associated cost function $c(e_i)=e_i^2$.

\subsubsection{Data-generating Process (for Application Instance)}

\paragraph{Tasks.} To show the effectiveness of Spot Check Equivalence, we consider an information elicitation application with $|J|=500$ tasks. To compare the Spot Check Equivalence of different performance measurements in various contexts, we vary $|J|$ according to Table~\ref{tab:ntasks}.

\paragraph{Assignment graph.} The data-generating process $D$ will randomly generate a bipartite graph between $I$ and $J$: To show the effectiveness of Spot Check Equivalence, each agent is assigned to $50$ tasks, while each task is assigned to $5$ agents. To compare the Spot Check Equivalence of different performance measurements in various contexts, we vary the parameters according to Table~\ref{tab:ntasks}.

\begin{table}[!ht]
    \centering
\begin{tabular}{cccc}
\toprule
$\#$ agents & $\#$ tasks & $\#$ tasks per agent & $\#$ agents per task \\
\midrule
50 & $50$ & $5$ & 5 \\
50 & $100$ & $10$ & 5 \\
50 & $50\times K$ & $5\times K$ & 5 \\
\bottomrule
\end{tabular}

where $K\in \{3,4,5,6,7,8,9,10\}$
    \caption{IEC parameter setups for ABM simulations}
    \label{tab:ntasks}
\end{table}

\paragraph{Ground truth and signals.} For generating the ground truth and signals, we will apply the generalized Dawid-Skene model from previous work \citep{dawid1979maximum,zhang2022high}, in which:

Each task $j\in J$ has a ground truth $g_j\in \mathcal{GT}$ where $\mathcal{GT}$ is a finite set including all possible ground truths. Since $\mathcal{GT}$ is finite, we can use a vector to represent the prior distribution of the ground truth $\omega$, whose $i$th item represents the probability of $i$th possible ground truth. For convenience, we also denote that vector as $\omega$.

    In our experiment, we use the $\omega$ learned from a dataset of a crowdsourcing task on Amazon Mechanical Turk \citep{zhang2022high,soton376543}. 
    \[
        \omega \approx [0.196, 0.241, 0.247, 0.316]
    \]

The agent $i$ will receive a signal $o_{ij}\in \Sigma$ on task $j$ given her effort level and the task's ground truth. In our experiment, we assume that $\Sigma = \mathcal{GT}$. Then, we can define two $|\mathcal{GT}|\times|\Sigma|$ confusion matrices, $\Gamma_{work}$ and $\Gamma_{shirk}$. The $(row,col)$ entry of $\Gamma_{work}$ and $\Gamma_{shirk}$ represents the probability of getting the $col$-th signal conditioned on the $row$-th ground truth when the agent exerts effort level 1 and 0 respectively.

When the agent $i$ exert $e_i$ effort, the confusion matrix is 
\[\Gamma_i = e_i * \Gamma_{work} + (1-e_i) * \Gamma_{shirk}\] 
where the confusion matrix $\Gamma_{work}$ is also learned from the above dataset \citep{zhang2022high,soton376543}, and we set $\Gamma_{shirk}$ as a matrix where each row represents a uniform distribution.
\[
\Gamma_{work} \approx 
\begin{bmatrix}
0.771& 0.122& 0.084& 0.024\\
0.091& 0.735& 0.130& 0.044\\
0.033& 0.062& 0.866& 0.039\\
0.068& 0.164& 0.099& 0.669
\end{bmatrix}
\]

\subsubsection{Performance Measurement}\label{subsec:performance-measurement}  We consider several performance measurements $M$ to evaluate the effectiveness of the Spot Check Equivalence. We implement several peer prediction mechanisms, which yield different Spot Check Equivalences within the above information elicitation context. 

\paragraph{Output Agreement (OA) \citep{von2004labeling,von2008designing,waggoner2014output}} In the OA mechanism, we pair the agents who work on the same task, and compare their reports. If the reports match, both agents receive a score of 1, and if they don't, the score is 0. One agent's total score is the summation of the scores on all the tasks assigned to him.

\paragraph{Peer Truth Serum (PTS) \citep{faltings2017peer}} Similar to OA, we pair the agents who work on the same task, and compare their reports. Instead of scoring 1 for agreement, we score the inverse of the frequency of the agent's report, which means, intuitively, we reward more for the agreement of the uncommon reports.

\paragraph{Correlated Agreement (CA) \citep{shnayder2016informed}} The CA mechanism is an extension of \citet{dasgupta2013crowdsourced}'s mechanism for multi-signal information elicitation. Without loss of generality, we denote the signal (report) space as $\Sigma = [1,2,...,n]$. Given two agents 1 and 2, let $r_1$ and $r_2$ denote the random variables of agent 1 and 2's report on all their tasks respectively. Then, we define an $n \times n$ delta matrix $\Delta$ as
\[
\Delta_{row,col} = \Pr[r_{1} = row, r_{2} = col] - \Pr[r_{1} = row]\Pr[r_{2} = col]
\]

We score the agent according to the agreement on positively correlated reports, i.e. we score agent 1 and agent 2 the following value. \[\sum_{row,col} \Delta_{row,col} \cdot \mathbbm{1} [\Delta_{row,col}>0]\]

\paragraph{$f$-Mutual Information ($f$-MI) \citep{kong2018water,kong2019information}} The $f$-MI mechanism scores the agent with the mutual information between her and her peer's reports, the intuition is that any manipulation on the two random variables will decrease the mutual information between them. We define the $f$-Mutual Information between agent 1 and 2's report as:
\[
f\text{-MI} = \sum_{row,col} \Pr[r_{1} = row, r_{2} = col] f\left(\frac{\Pr[r_{1} = row]\Pr[r_{2} = col]}{\Pr[r_{1} = row, r_{2} = col]}\right)
\]
where $f$ is a convex function and $f(1)=0$. For example, when we take $f(x)=-\log (x)$, then $f$-MI is the Shannon Mutual Information. Note that the $f$-Mutual Information could be regarded as the $f$-divergence ($D_{f} = \sum_x p(x) f \left(\frac{q(x)}{p(x)}\right)$) \citep{ali1966general} of the joint distribution and the product of the marginal distributions of the agents' reports. 

We list all the $f$-Mutual Informations we will use in our simulations in Table~\ref{tab:f-mi}. Note that when applying a Total Variation Distance divergence, the $f$-MI mechanism is almost equivalent to the CA mechanism, thus, we only show the result of the CA mechanism in our simulations.

\begin{table}[]
    \centering
\renewcommand{\arraystretch}{1.2}
\begin{tabular}{ccc}
\toprule
$f(x)$ & $f$-divergence & Notation \\
\midrule
$-log(x)$ & KL-divergence & KL \\
$|x-1|$ & Total Variation Distance & TVD \\
$(\sqrt{x}-1)^2$ & Squared Hellinger & $H^2$ \\
\bottomrule
\end{tabular}
    \caption{$f$-Mutual Information}
    \label{tab:f-mi}
\end{table}

\paragraph{Determinant-based Mutual Information (DMI) \citep{kong2020dominantly}} \citet{kong2020dominantly} generalize the Shannon Mutual Information to the Determinant-based Mutual Information (DMI). Let an $n\times n$ matrix $\mathbf{U}_{X,Y}$ denote the joint distribution of two random variables, the DMI of these two random variables are defined as $DMI(X;Y) = |\det(\mathbf{U}_{X,Y})|$. 

Then, given two agents 1 and 2, the score of the DMI mechanism is defined as the product of the DMI of the two agents' reports on two disjoint partitions of all the tasks. Specifically, we divide all the common tasks of 1 and 2 into two parts, calculate their DMI in these two parts respectively, and score each agent with the product of the two DMI.

However, even though DMI demonstrates impressive theoretical properties, it does not perform well in our simulation due to the considerable noise of its performance score (Figure~\ref{fig:gr-effort}). 

\paragraph{Spot-checking}
In addition, to calculate the Spot Check Equivalence, we also implement a spot-checking performance measurement (Definition~\ref{def:sc}) as a benchmark, in which we use accuracy as the quality function, i.e. $Q(r,g) = \mathbbm{1}[r=g]$. We denote this spot-checking performance measurement as $SC(X)$ where $X$ is the checking percentage.

\subsubsection{Payment Scheme}

We now introduce the tournament payment scheme we use in our simulation.

\paragraph{Borda-count payment scheme.} A very intuitive way to pay an agent according to her ranking is to pay her how many agents she beats. When there is a draw, we split the payoff evenly. Formally, we have 
\[p_i = C\cdot \#\text{beaten}= C \sum_{i' \in |I|, i' \neq i} \left( \mathbbm{1} [s_i > s_{i'}] + \frac{1}{2} \mathbbm{1} [s_i = s_{i'}]\right)\]
where $C$ is a constant parameter and the total payment is $C\times \binom{|I|}{2}$.

To calculate the total payment\footnote{Note that the total payment of Borda-count is deterministic, thus, we use ``total payment'' in the rest of this section instead of ``expected total payment''.} in the Borda-count scheme for a specific performance measurement $M$ for the symmetric local equilibrium where every agent exerts $e_i=\xi$ effort. We let the derivative of the agent $i$'s expected payoff equal to $0$, which implies the parameter $C$ should be
\[
C = \frac{\frac{\partial}{\partial e_i}\E[\#\text{beaten}|e_i,e_{-i}=\xi]|_{e_i=\xi}}{\frac{\partial}{\partial e_i}c(e_i)|_{e_i=\xi}}
\]

Note that to guarantee Individual Rationality (The agents' expected utility is non-negative), when the calculated optimal payment is less than the total cost of all the agents, we set the total payment as the total cost. We assume that if a payment scheme can incentivize effort level $\xi$ using the optimal payment, it can also incentivize the same effort when the total payment is greater than the optimal payment.

\subsection{Measurement-Integrity-based Spot Check Equivalence.}\label{subsec:MI-SEC}

We examine whether the Spot Check Equivalence based on Measurement Integrity can indicate a performance measurement's motivational proficiency. Recall that the motivational proficiency of a performance measurement could be quantified by the amount of the expected total payment we need to elicit a fixed effort level in an information elicitation context. Thus, in the experiment examining the effectiveness, we mainly investigate the relationship between the Measurement Integrity and the expected total payment.

For several fixed effort levels, we apply all the performance measurements and estimate their Measurement Integrity and the total payment needed to elicit that equilibrium with the Borda-count payment scheme by iterating the data-generating process 5000 times. Detailed implementation is shown in Appendix~\ref{apx:abm}.

Recall that to satisfy Individual Rationality, the total payment needs to be at least the agents' cost to exert the effort. Since the minimal payment is the same for all performance measurements when the effort level is the same, we visualize that as a horizontal line in our result.

\begin{figure}[!ht]
  \centering
  \subfigure[effort level 0.5]{
    \includegraphics[width=0.47\linewidth]{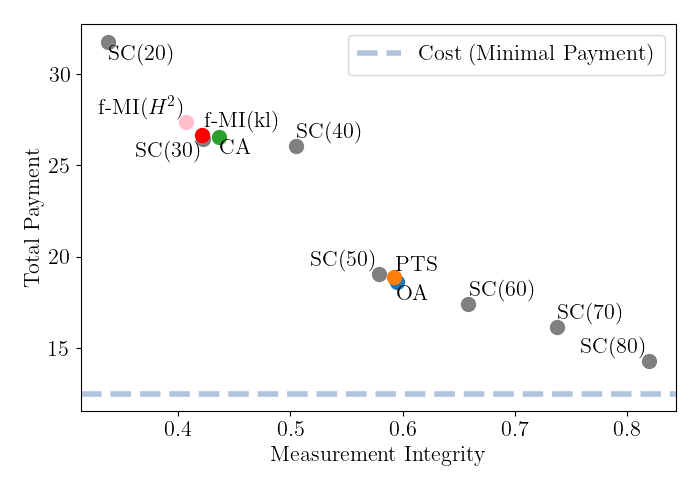}
  }
  \subfigure[effort level 0.6]{
    \includegraphics[width=0.47\linewidth]{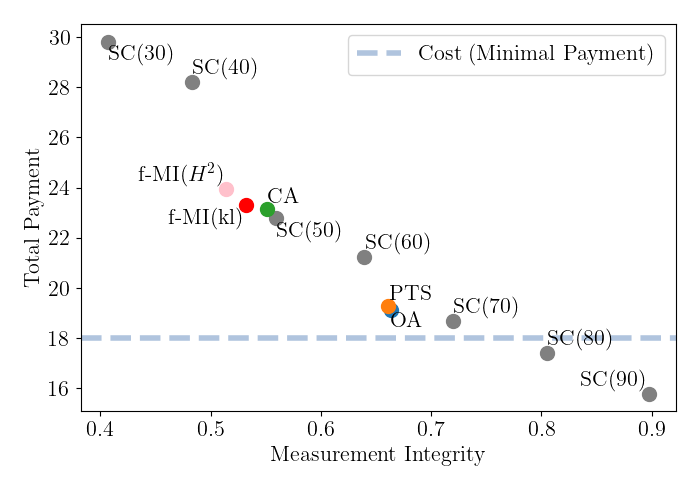}
  }
  \subfigure[effort level 0.7]{
    \includegraphics[width=0.47\linewidth]{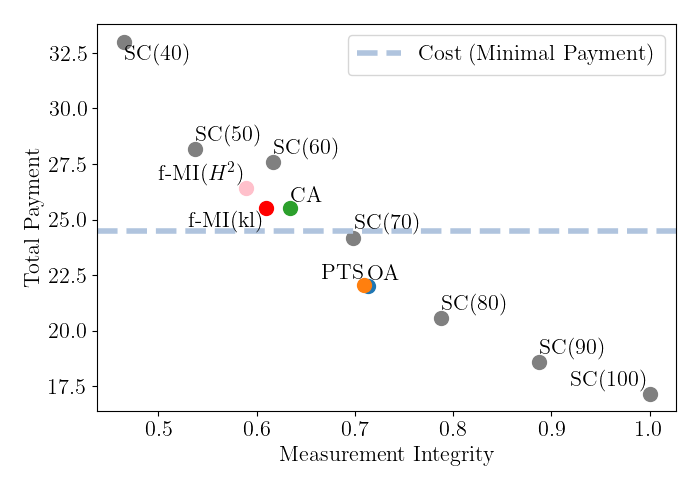}
  }
  \subfigure[effort level 0.8]{
    \includegraphics[width=0.47\linewidth]{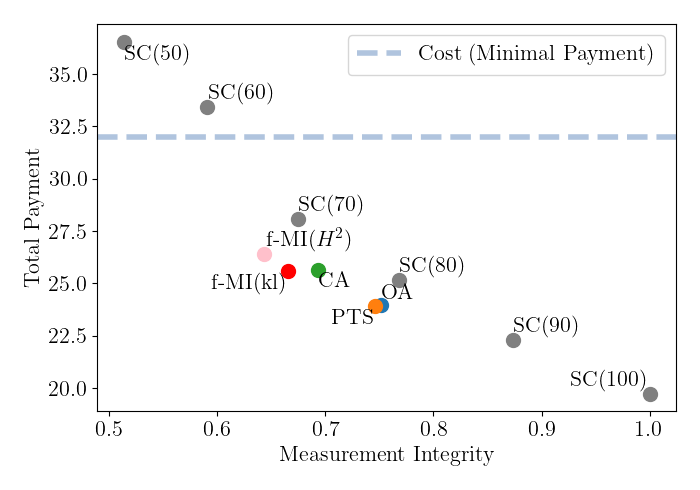}
  }
  \caption{Measurement Integrity v.s Total Payment of Borda-count payment scheme: the $x$-axis is the Measurement Integrity and the $y$-axis is the total payment needed to elicit that equilibrium within the tournament payment scheme. The horizontal line shows the agents' cost to exert the effort level, which implies the minimal payment to satisfy Individual Rationality.}
  \label{fig:effective-gr}
\end{figure}

With the results in Figure~\ref{fig:effective-gr}, we can observe that:
\begin{enumerate}
    \item The Measurement Integrity monotonically increases with the spot-checking ratio.
    \item The Measurement Integrity and the total payment are significantly negatively correlated.
\end{enumerate}

This implies that, at a symmetric equilibrium, if a performance measurement $M$ has Measurement Integrity equal to spot-checking $SCE\%$, then it has a similar motivational proficiency, i.e. similar total payment, to that spot-checking performance measurement within a tournament payment scheme. Therefore, a higher Spot Check Equivalence indicates higher motivational proficiency. 

Note that, even if considering IR, the motivational proficiency is still monotonically increasing with the Spot Check Equivalence, however, when IR is binding, more spot-checking does not further decrease the total payment.

\subsubsection{Measurement Integrity is a computationally efficient proxy.}

\begin{figure}[!ht]
  \centering
  \subfigure[Measurement Integrity]{
    \includegraphics[width=0.45\linewidth]{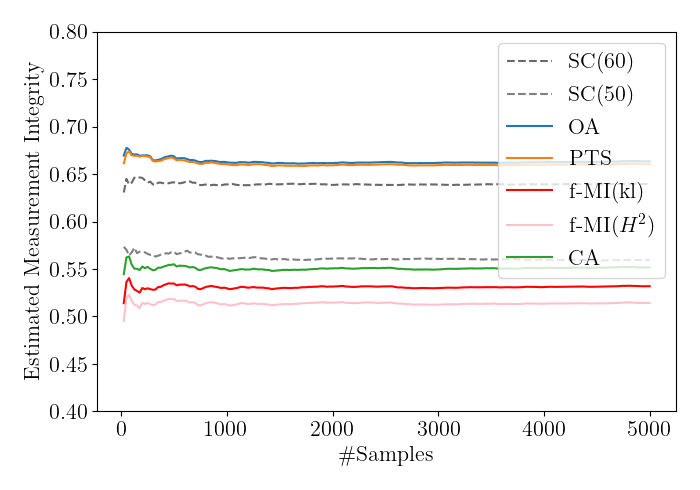}
  }
  \subfigure[Total payment of Borda-count]{
    \includegraphics[width=0.45\linewidth]{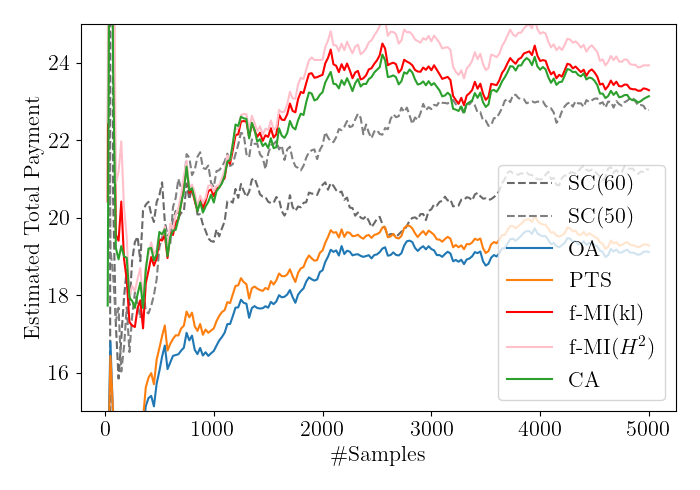}
  }
  \caption{Convergence speed of the Measurement Integrity and the total payment: the $x$-axis is the number of the samples, and the $y$-axis is the estimated Measurement Integrity and the estimated total payment of the Borda-count payment scheme at effort level $\xi=0.6$ respectively.}
  \label{fig:convergence}
\end{figure}

In addition, we find that the Measurement Integrity converges significantly faster than the total payment. Figure~\ref{fig:convergence} illustrates the variation in both the Measurement Integrity and total payment as the number of iterations goes up.

In the previous results of Figure~\ref{fig:effective-gr} (b), we find that when eliciting effort level of $\xi=0.6$, the $f$-MI(kl), $f$-MI($H^2$) and CA performance measurements have a little less motivational proficiency comparable to $50\%$ spot-checking. Meanwhile, the OA and PTS performance measurements are better than $60\%$ spot-checking.

Figure~\ref{fig:convergence} demonstrates that achieving the same outcome requires significantly fewer iterations for the calculation of Measurement Integrity compared to the total payment. This suggests that even when it's possible to compute the expected total payment (e.g. with agent-based model simulation), utilizing Measurement Integrity as a proxy offers better computational efficiency.

\subsection{Sensitivity-based Spot Check Equivalence}\label{subsec:exp-sensitivity}
\begin{figure}[!ht]
  \centering
  \subfigure[effort level 0.5]{
    \includegraphics[width=0.47\linewidth]{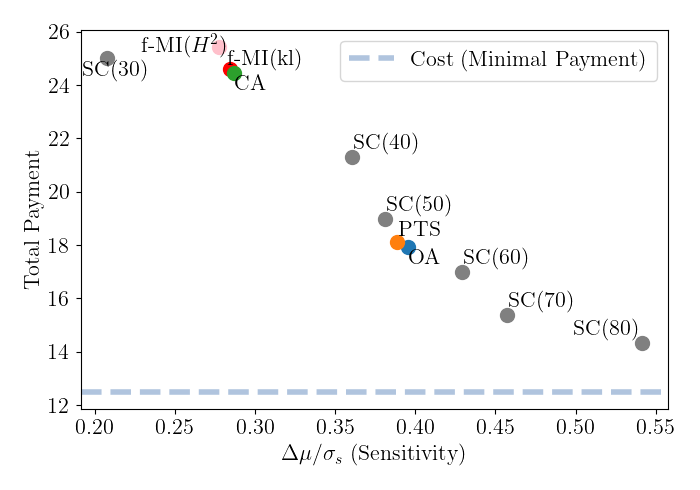}
  }
  \subfigure[effort level 0.6]{
    \includegraphics[width=0.47\linewidth]{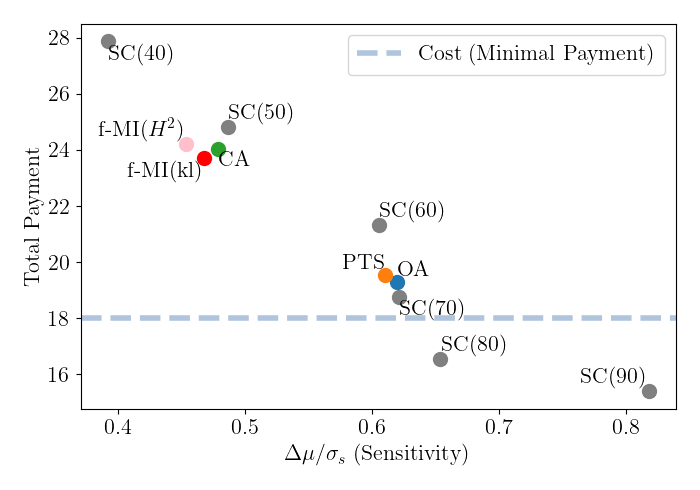}
  }
  \subfigure[effort level 0.7]{
    \includegraphics[width=0.47\linewidth]{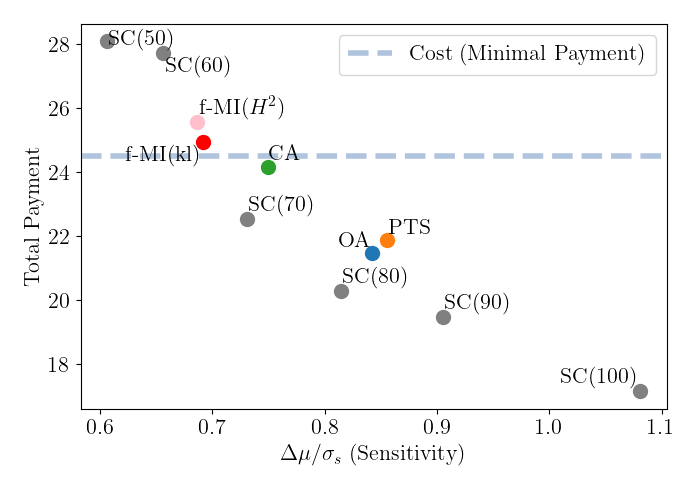}
  }
  \subfigure[effort level 0.8]{
    \includegraphics[width=0.47\linewidth]{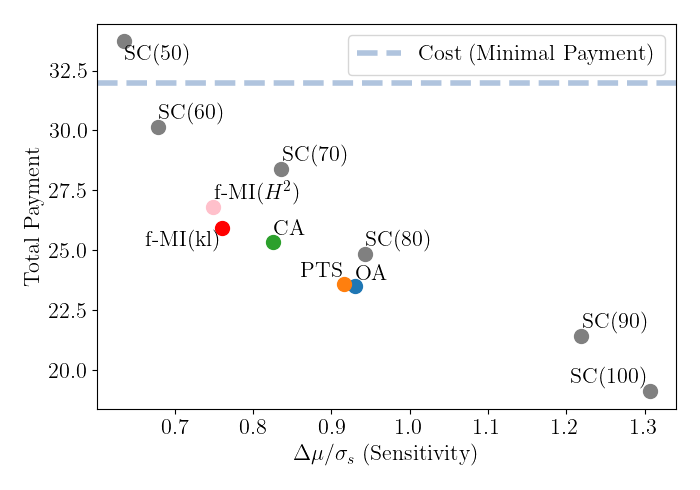}
  }
  \caption{Sensitivity v.s Total Payment of Borda-count payment scheme: the $x$-axis is the $\Delta\mu/\sigma$ and the $y$-axis is the total payment needed to elicit that equilibrium within the tournament payment scheme. The horizontal line shows the agents' cost to exert the effort level, which implies the minimal payment to satisfy the Individual Rationality.}
  \label{fig:effective-gr-no-gt}
\end{figure}

We now examine the effectiveness of the Spot Check Equivalence based on Sensitivity as a metric of motivational proficiency. Here, when estimating $\Delta\mu/\sigma$ (which is proportional to the Sensitivity), we consider a more realistic scenario where there is only one sample from the data generating process and no ground truth. We estimate $\Delta\mu/\sigma$ for each performance measurement according to Algorithm~\ref{alg:sce_no_gt} with $T=5000$ iterations. We then compare the $\Delta\mu/\sigma$ with the total payment estimated in the same way as in the previous subsection. The results are shown in Figure~\ref{fig:effective-gr-no-gt}.

Similarly, with the results in Figure~\ref{fig:effective-gr-no-gt}, we can observe that: (1) The Sensitivity monotonically increases with the spot-checking ratio. (2) The Sensitivity and the total payment are significantly negatively correlated. This observation implies that the Spot Check Equivalence based on Sensitivity can be used as a metric of motivational proficiency 

\subsection{Spot Check Equivalence of Peer Prediction}

\subsubsection{Peer Prediction works better to elicit high effort.}

We apply the workflow in our agent-based model experiment to further compare the Spot Check Equivalence of different performance measurements in various contexts. Previous works \citep{zhang2022high,burrell2021measurement,gao2016incentivizing}  conduct comparisons of certain metrics in specific contexts. To further study the motivational proficiency of the performance measurements, we calculate the Spot Check Equivalence across various information elicitation contexts.

\begin{figure}[!ht]
  \centering
    \includegraphics[width=0.5\linewidth]{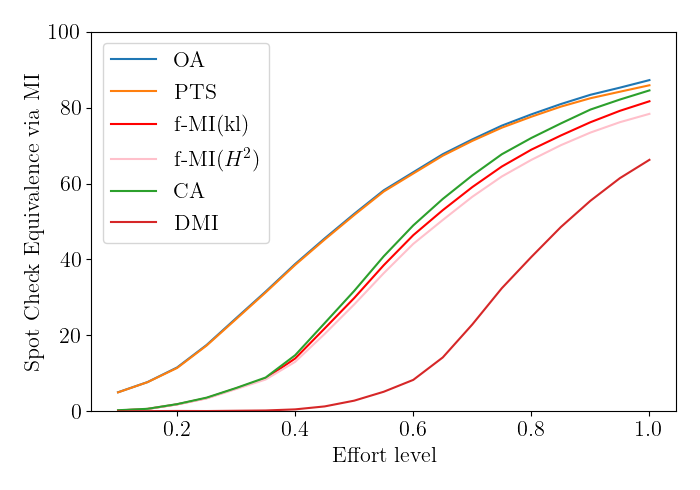}
  \caption{Spot Check Equivalence on different effort levels}
  \label{fig:gr-effort}
\end{figure}

We then enumerate the effort levels and calculate the Spot Check Equivalence via the Measurement Integrity of each performance measurement (Figure~\ref{fig:gr-effort}).

We find that when eliciting a low-effort equilibrium the Spot-Checking Equivalence of peer prediction performance metrics is low, however, the relative motivational proficiency of peer prediction increases fast. That is because a peer prediction performance measurement scores an agent according to the correlation between her report and her peers', when every agent exerts a low effort, her peers' reports are noisy so the score is quite noisy.

\subsubsection{Mutual-information-based mechanisms work better when $\#$tasks per agent increases.}

As the number of tasks per agent increases, the SCE of OA and PTS remains the same, while for the mutual-information-based mechanisms (CA, $f$-MI), the Spot Check Equivalence significantly increases. As the $\#$tasks per agent increases, the estimation of the joint distribution of two agents' reports will be more accurate, which leads to a more accurate score.

\begin{figure}[!ht]
  \centering
  \subfigure[effort level 0.5]{
    \includegraphics[width=0.47\linewidth]{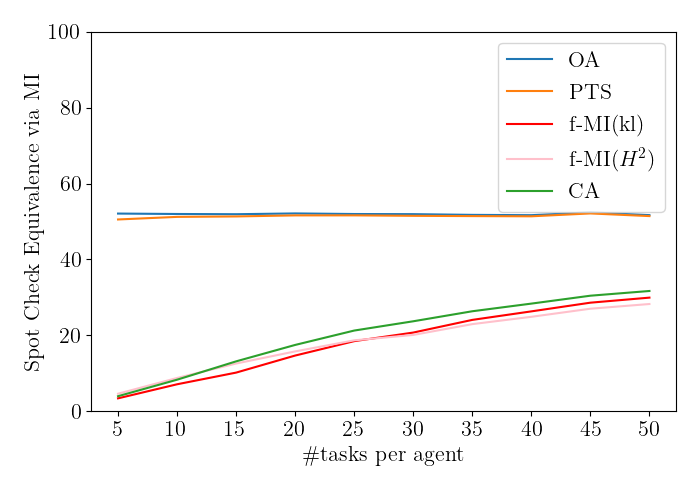}
  }
  \subfigure[effort level 0.7]{
    \includegraphics[width=0.47\linewidth]{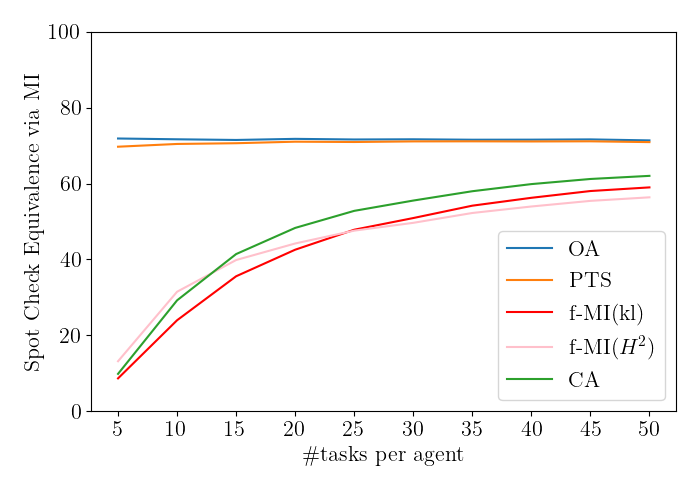}
  }
  \subfigure[effort level 0.9]{
    \includegraphics[width=0.47\linewidth]{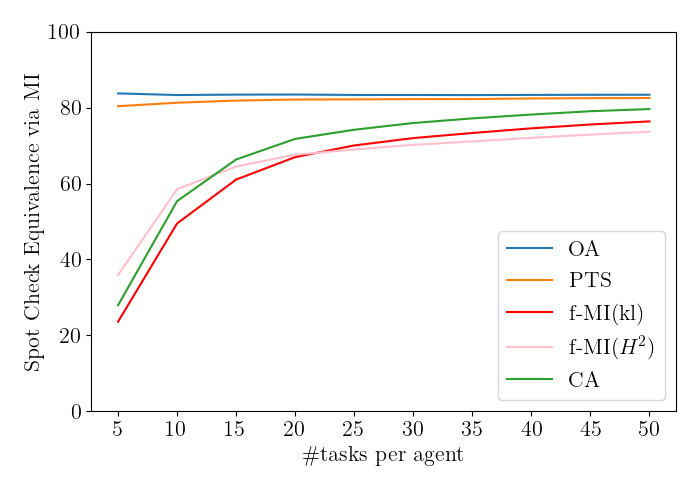}
  }
  \caption{\#tasks per agent vs. the Spot Check Equivalence based on Measurement Integrity: the $x$-axis is the number of tasks assigned to each agent and the $y$-axis is the Spot Check Equivalence calculated by Measurement Integrity.}
  \label{fig:rq2_mi}
\end{figure}

\section{Does Peer Prediction Make Things Worse?}\label{sec:dis}

In this section, we provide a detailed discussion about why \citet{gao2016incentivizing, GAO2019618} have the result that ``peer prediction makes things worse'' which contradicts the other literature \cite{zhang2022high} and our main results that show peer-prediction mechanisms can have non-zero Spot Check Equivalence.

We first introduce the Information Elicitation Context in their paper.

\paragraph{Agent} $Agent=(I,c,\mathbf{e})$. The agent can choose a binary effort $e\in \{0,1\}$, where exerting high effort has a cost $c(1)=c^{E}$ and exerting no effort has a cost $c(0)=0$. 

\paragraph{Application Abstraction} In $App=(J,\mathcal{GT},\omega,\Sigma,D)$, they also model signal generation as a random function: 
\[D_{signal}: \{0,1\}\times  \mathcal{GT} \rightarrow \Delta\Sigma\]
however, importantly, the signals of agents are not i.i.d. across agents.  When the agent exerts high effort, she will get a high-quality signal, which is drawn from a distribution conditional on the task's ground truth.  When the agent exerts no effort, she will get a low-quality signal that is uncorrelated with the task's ground truth, but, crucially, the no-effort signals are perfectly correlated across no-effort agents---they all receive the same signal. 

\paragraph{Performance Measurement}
Firstly, they assume that the scoring function in the spot-checking performance measurement (Definition~\ref{def:sc}) can effectively incentivize high effort, i.e.
\[\E\left[S\left(D_{signal}(1,g_j),g_j\right)\right] - c^E > \E\left[S\left(D_{signal}(0,g_j),g_j\right)\right]\]

In addition to the spot-checking performance measurement (Definition~\ref{def:sc}), they propose a spot-checking peer-prediction performance measurement, where for the unchecked tasks, they apply a peer-prediction performance measurement to score the agents.

\paragraph{Payment Scheme} They fix a function $f:\mathcal{GT} \times \Sigma \rightarrow \mathbb{R}^*$.   The payment scheme is additive across tasks and pays agents according to $f$ for answering spot-checked tasks, and according to a peer-prediction mechanism for tasks that are not spot-checked. 

\paragraph{Agent reports} They allow the agents to strategically report their signals.

\paragraph{Equilibrium} They focus on two possible equilibria. In the no-effort equilibrium, each agent exerts no effort and uses the same strategy to report her signal. In the truthful equilibrium, each agent exerts high effort and truthfully reports her signal.

\subsection{Comparing Spot-checking Mechanisms and Spot-checking Peer-prediction Mechanisms}

Notice that, by assumption, with enough spot-checking, you will get the truthful equilibrium. They use the minimum spot-checking ratio that ensures the truthful strategy profile is an equilibrium as a measure of the mechanisms' performance. Formally, they use $p_{Pareto}$ to denote the minimum spot-checking ratio where the truthful equilibrium Pareto dominates the no-effort equilibrium when applying the hybrid peer-prediction/spot-checking mechanism.  They use $p_{ds}$ to denote the minimum spot-checking ratio where the truthful strategy profile is a dominant strategy in spot-checking mechanism.

Then they propose a theorem for comparing the spot-checking mechanism and the spot-checking peer-prediction mechanism:

\begin{theorem} [Section 5 Theorem 3 in \citet{gao2016incentivizing}]\label{thm:pp_makes_things_worse}
    For any spot-checking peer-prediction mechanism, if the no-effort equilibrium exists and Pareto dominates the truthful equilibrium when the cost of effort is 0 ($c^E = 0$) and no task is checked ($p = 0$), then $p_{Pareto}\geq p_{ds}$ for any  $c^E \geq 0$.
\end{theorem}

There are three differences between the models in \citet{gao2016incentivizing} and \citet{zhang2022high}  that account for this stark difference.

The most obvious difference is that the payments in \citet{gao2016incentivizing} are restricted to be additive across tasks.  This is rather similar to assuming a linear payment rule when the score is additive across tasks.  However, \citet{zhang2022high}  use tournaments, which are not linear and typically have much better motivational proficiency.   

Second, \citet{gao2016incentivizing}'s assumption of a fixed payment function $f$ is extremely restrictive.   As we discussed in Section~\ref{sec:intro}, when the spot-checking ratio decreases, it is possible to maintain the same incentive properties by simply scaling up the payment function.  Thus, if we scale up $f$, we can make $p_{ds}$ arbitrarily small, and, conversely, by scaling down $f$, we can make $p_{ds}$ arbitrarily close to 1.  
On the other hand, in \citet{zhang2022high} the mechanism is defined as a performance measurement and a payment scheme, and the payment scheme is optimized to work with the scoring function, rather than being artificially fixed.  

Finally, \citet{gao2016incentivizing}' make a key assumption on the no-effort signals being perfectly correlated. For example, in peer grading, the writing and formatting quality is a signal that can be accessed with very little effort while assessing the correctness both requires more effort and will likely lead to less agreement. 

Notice that the premise of Theorem~\ref{thm:pp_makes_things_worse} is that when the cost of effort is 0 ($c^E = 0$) and no task is checked ($p = 0$), the equilibrium where agents exert no effort Pareto dominates the truthful equilibrium.

Let's zoom in on this.  First, consider the case where the cost of the high effort signal $c^E=0$.  Notice that any spot-checking mechanism that checks any positive ratio of tasks will have the high-effort profile as an equilibrium because agents will receive some positive payoff, but have 0 cost. 

Next, consider a peer-prediction mechanism (e.g. a hybrid peer-prediction/spot-checking mechanism with spot-checking ratio 0).  Again, the theorem is vacuously true, unless the profile in which no agent exerts effort Pareto dominates the high-effort equilibrium because, in the former, all agents agree and receive a maximal payoff\footnote{They are assuming here that complete agreement brings a maximum payoff.}, but, in the latter, they do not all agree.  

Together, this shows that any peer prediction mechanisms will have a Spot Check Equivalence of $0$, since in this case, any spot-checking ratio that is greater than 0 will inevitably lead to the truthful equilibrium given that the effort cost is nonexistent. 

Because having a Spot Check Equivalence of $0$ is an assumption of the theorem, it is no wonder that such peer-prediction mechanisms do not help!

However, this still potentially makes for a very strong critique of peer-prediction mechanisms because in many real-world settings, there exist cheap signals.  For example, as mentioned in the introduction, when humans are labeling LLM responses, it is much easier to judge them on how authoritative-sounding the responses are than on how truthful the responses are.  However, such labels may encourage hallucinations. 

Indeed, \citet{gao2016incentivizing} led to several papers trying to create peer-prediction mechanisms that are robust against ``cheap'' signals  (i.e. the no-effort/low-effort signals that can bring higher agreement than the high-effort signals).

\citet{kong2018eliciting} propose a peer prediction mechanism called Hierarchical Mutual Information Paradigm (HMIP), assuming a hierarchical information structure where high-effort (or higher expertise) agents have access to the information of low-effort (or lower expertise) agents. HMIP encourages agents to invest effort and incentivizes truthful reporting by paying the high-effort agents for correctly predicting the ``cheap'' signals from the low-effort agents. 
Additionally, a human subject experiment \citep{kong2022eliciting}  shows evidence of a hierarchical information structure among the participants.

\section{Related Work} \label{sec:related}

Besides the theoretical literature discussing peer prediction mechanisms, as highlighted in Section~\ref{sec:exp}, there are empirical studies that validate these mechanisms. For instance, \citet{radanovic2016incentives} experimentally tested their peer prediction mechanism in both peer grading and crowdsourcing scenarios to validate its theoretical properties. Similarly, \citet{shnayder2016informed} employed peer grading data from the edX MOOC platform to assess the performance of their proposed mechanisms. Spot checking in peer grading scenarios has already been empirically examined in works such as \cite{zarkoob2019report,wright2015mechanical}. Additionally, \citet{goel2019deep} study combining peer prediction and spot-checking, and introduce the Deep Bayesian Trust Mechanism that utilizes peer reports to reduce the need for spot-checking.

Furthermore, in forecasting contexts where agents are rewarded afterward based on the agreement between their forecasts and the outcomes, i.e. the ground truth is accessible for little or no cost, \citet{hartline2022optimal,li2022optimization,neyman2021binary} study how to optimize proper scoring rules to incentivize effort, and consequently, elicit high-quality information. These works suggest the possibility of optimizing the spot-checking mechanisms by scoring the checked tasks according to the optimal proper score rules, which indicates possible direction for future research.

\subsection{Discussion of the truthful report assumption.}\label{subsec:truthful-asp} In our Section~\ref{sec:model}, we assume that the agents will truthfully report their signal. This assumption is reasonable when applying a linear payment scheme given the performance measurement is truthful under certain settings.
In particular, \citet{dasgupta2013crowdsourced} propose the first multi-task peer prediction mechanism. Later on, the CA mechanism \cite{shnayder2016informed} and $f$-MI mechanism \cite{kong2019information} are proposed to generalize \citet{dasgupta2013crowdsourced}'s mechanism from binary signal space to finite signal space, the MA mechanism \cite{10.1145/3543507.3583292} generalizes CA to address task-dependent strategies,  the DMI mechanism \cite{DMI_Kong} aims to reduce the required number of tasks, \citet{schoenebeck2020learning} handle continuous signals with a learning-based mechanism, \citet{Agarwal2017-ty} deal with the heterogeneous agents, and \citet{rowdycrowds} address adversarial attacks in peer prediction.

Furthermore, \citet{burrell2021measurement} examine the robustness of the truth-telling equilibrium with agent-based model experiments. However, when applying a non-linear payment scheme, e.g. winner-take-all, the agents may have the incentive to strategically report their signal, e.g. increasing the variance of their score to get a higher probability of being the winner. \citet{zhang2022high} propose a truthful winner-take-all payment scheme by adding noise to the agents' score which may hurt the incentive for effort. However, further study needs to be conducted to study the robustness of different performance measurements against strategic reports with other non-linear payment schemes. This gap indicates another potential future direction of our research.

\section{Conclusion and Discussion} \label{sec:conclusion}

In summary, our research provides a methodology for understanding the performance, especially motivational proficiency, of information elicitation mechanisms in various contexts, the Spot Check Equivalence, and consequently offers valuable insights for the design of effective and efficient incentive mechanisms that promote the acquisition of high-quality information.

Future research might be conducted to investigate motivational proficiency in a non-monetary setting, e.g. in peer grading, we care about how to elicit agents' effort with the bounded individual payoff, since the students' grades could only be A, B, C, F, etc. Another future direction might be to study motivational proficiency in a more sophisticated model where the agents have heterogeneous cost functions. 

\section*{Acknowledgments}
We would like to thank Noah Burrell for his invaluable help and profound discussions that are instrumental in initiating this paper. 

\newpage

\bibliographystyle{ACM-Reference-Format}
\bibliography{ref}

\newpage

\appendix

\section{Proof of Theorem~\ref{thm:main}}\label{apx:theory}

\newtheorem*{thm-shared}{Theorem~\ref{thm:main}}

\begin{thm-shared}[Main Theorem] \label{thm:unification}
For a given performance measurement $M$ within an information elicitation context $IEC$ where every agent exerts effort level $\xi$, when Assumption~\ref{asp:gaussian-quality} \ref{asp:gaussian-score} and \ref{asp:score-independence} are satisfied, there exists a linear bijection between the ${\operatorname{ MI }}_{Q, \operatorname{corr}}(IEC\leftarrow M)$ and the Sensitivity $\delta(\xi)$, where $\operatorname{corr}$ is the sample Pearson correlation coefficient and the number of agents goes to infinity.
\end{thm-shared}

\begin{proof}

Recall the definition of Measurement Integrity, we have that 

\[\underset{Q, \operatorname{corr}}{\text { MI }}(IEC\leftarrow M)=\mathbb{E}_{IEC}\left[\operatorname{corr}\left(\mathbf{s}, \mathbf{q}\right)\right]\]
where $\mathbf{s}, \mathbf{q}$ represents the vector of all agents' performance scores and report qualities respectively.

In the above definition, we apply the \textit{sample Pearson correlation coefficient} to $\operatorname{corr}$, i.e.,
\[
\operatorname{corr}(\mathbf{s}, \mathbf{q})=r(\mathbf{s}, \mathbf{q})=\frac{\sum_{i}\left(s_i-\bar{s}\right)\left(q_i-\bar{q}\right)}{\sqrt{\sum_{i}\left(s_i-\bar{s}\right)^2} \sqrt{\sum_{i}\left(q_i-\bar{q}\right)^2}}
\]
where $\bar{s}=\frac{1}{|I|} \sum_{i} s_i$ (the average performance score); and analogously for $\bar{q}$. 

Recall that, by Assumption~\ref{asp:score-independence}, we have both that, when the number of agents is large, the agents' qualities are independent and the performance scores of the agents are independent. Thus, we can regard each agent's quality and performance score pair as a sample from a joint distribution. In addition, the sample correlation coefficient $r$ is a consistent estimator of the \textit{population Pearson correlation coefficient} $\rho$ as the sample size grows large, which is defined as
\[\operatorname{corr}(s, q)=\rho(s, q)=\frac{COV(s,q)}{\sigma_s \cdot \sigma_q}\]
where $s,q$ are random variables representing the score and quality of one agent respectively.

Therefore, we have the following proposition.

\begin{proposition}
\[
     \lim_{|I|\rightarrow +\infty}\E[r(\mathbf{s},\mathbf{q})] = \rho(s,q)
\]
where $s,q$ are random variables representing the score and quality of one agent respectively.
\end{proposition}

We then show that there exists a bijection between $\rho(s,q)$ and $\delta(\xi)$. Recall the definition of Sensitivity \[\delta(\xi) = \frac{\frac{\partial }{\partial e_i}\mu_{s}(e_i)|_{e_i=\xi}}{\sigma_s(\xi)}\]

We can find that there is $\frac{1}{\sigma_s(\xi)}$ in both the Sensitivity $\delta(\xi)$ and correlation coefficient $\rho(s,q)$, so we only need to find a bijection between $\frac{COV(s,q)}{\sigma_q}$ and $\frac{\partial }{\partial e_i}\mu_{s}(e_i)|_{e_i=\xi}$.

In the following discussion, since bias does not affect the correlation coefficient, we can assume $\xi=0$ without loss of generality. 

\subsubsection{From the MI side}

Here, every probability is conditioned on $e=\xi$.
\begin{align*}
    COV(s,q) = & \E[s\cdot q] - \E[s]\cdot \E[q] \\
    = & \int_{q,s} s \cdot q \cdot \Pr[s,q] \operatorname{ds} \operatorname{dq} \tag{$\E[q]=\xi=0$}\\
    = & \int_{q,s} s \cdot q \cdot \Pr[s|q] \cdot Pr[q] \operatorname{ds} \operatorname{dq}\\
    = & \int_{q} q \cdot Pr[q] \left(\int_{s} s \cdot \Pr[s|q] \operatorname{ds} \right) \operatorname{dq} \\
    = & \int_{q} q \cdot Pr[q] \cdot \mu_{s|q} (q) \operatorname{dq}
\end{align*}

\subsubsection{From Sensitivity side}

\begin{align*}
     \frac{\partial}{\partial e}\mu_s(e) = & \frac{\partial}{\partial e}\int_q \mu_{s|q}(q)  \Pr[q|e] \operatorname{dq}  \\
     = & \int_q \mu_{s|q}(q) \frac{\partial}{\partial e} \Pr[q|e] \operatorname{dq} \\
\end{align*}

Here, since $q$ follows a normal distribution with mean $e$ and standard deviation $\sigma_q$, we have that
\begin{align*}
    \frac{\partial}{\partial e} \Pr[q|e] = & \frac{\partial}{\partial e} \frac{1}{\sqrt{2\pi} \sigma_q} \exp{-\frac{(q-e)^2}{2\sigma_q(e)^2}} \\
    = & \frac{1}{\sqrt{2\pi} \sigma_q(e)^3} \exp{-\frac{(q-e)^2}{2\sigma_q(e)^2}} \left((q-e) + \frac{(q-e)^2\sigma_q'(e)}{\sigma_q(e)}\right)\\
    = & \frac{1}{\sigma_q^2} \cdot q \cdot \Pr[q|e] 
    \tag{Assumption~\ref{asp:gaussian-quality}}
\end{align*}

Thus, by combining the above two equations, we have
\begin{align*}
    \frac{\partial}{\partial e}\mu_s(e) = & \int_q \mu_{s|q}(q) \cdot \frac{1}{\sigma_q^2} \cdot q \cdot \Pr[q|e] \operatorname{dq} \\
    = & \frac{1}{\sigma_q^2} \int_q q \cdot Pr[q|e] \cdot \mu_{s|q} (q) \operatorname{dq}
\end{align*}

Letting $e=\xi$, we have

\begin{align*}
    \frac{\partial}{\partial e}\mu_s(e)
    = & \frac{1}{\sigma_q^2} \int_q q \cdot Pr[q|e=\xi] \cdot \mu_{s|q} (q) \operatorname{dq} \\
    = & \frac{COV(s,q)}{\sigma_q^2}
\end{align*}

Therefore, we get the linear bijection between Sensitivity and correlation is \[\delta = \operatorname{corr}(s,q)/\sigma_q.\] 
\end{proof}






\section{Simulation Details}\label{apx:abm}

To show the effectiveness of the Measurement-Integrity-based Spot Check Equivalence, we can plug different performance measurements in the information elicitation context described in the previous subsection, then calculate the Measurement Integrity (which implies the Spot Check Equivalence) and the motivational proficiency (i.e. the total payment) respectively, and finally, observe the correlation between the Spot Check Equivalence and the motivational proficiency.

For a given information elicitation context and a target equilibrium effort level $\xi$, we propose the following workflow for one sample:

\begin{enumerate}
    \item Set all the agents' effort levels at $\xi$.
    \begin{enumerate}
        \item Generate a sample for the application abstraction according to the data-generating process $D$.
        \item Use the performance measurement $M$ to produce all the agents' scores $\mathbf{s}$.
        \item Use the quality function $Q$ to calculate the quality of the agents' report, and then calculate the Pearson correlation coefficient.
    \end{enumerate}
    \item Change agent 1's effort level to $e_1=\xi-\varepsilon$
    \begin{enumerate}
        \item Generate another sample for the application abstraction according to the data-generating process $D$.
        \item Use the performance measurement $M$ to produce all the agents' scores $\mathbf{s}'$.
    \end{enumerate}
\end{enumerate}

\paragraph{Calculating the Measurement Integrity.}
We sample 5000 times and use the mean of the Pearson correlation coefficient in the samples\footnote{It is a consistent estimator of the Measurement Integrity.} as the Measurement Integrity of performance measurement $M$. We apply the same workflow to all the performance measurements in Section~\ref{subsec:performance-measurement}, as well as spot-checking performance measurements with different checking ratios.

\paragraph{Calculating the total payment.}
We sample 5000 times and use the agents' scores $\mathbf{s}$ and $\mathbf{s}'$ to estimate the ${\frac{\partial}{\partial e_i}\E[\#\text{beaten}|e_i,e_{-i}=\xi]|_{e_i=\xi}}$, which can be used to compute the total payment for the Borda-count payment schemes.

\end{document}